# A Survey on Deep Reinforcement Learning-based Approaches for Adaptation and Generalization

**Pamul Yadav, Ashutosh Mishra, Junyong Lee and Shiho Kim**[*]
School of Integrated Technology, Yonsei University, South Korea
{pamul, ashutoshmishra, jjunilee, shiho}@yonsei.ac.kr
*Corresponding author: shiho@yonsei.ac.kr

## Abstract

Deep Reinforcement Learning (DRL) aims to create intelligent agents that can learn to solve complex problems efficiently in a real-world environment. Typically, two learning goals: *adaptation,* and *generalization* are used for baselining DRL algorithms' performance on different *tasks* and *domains*. This paper presents a survey on the recent developments in DRL-based approaches for *adaptation* and *generalization*. We begin by formulating these learning goals in the context of *task* and *domain*. Then we review the recent works under those approaches and discuss future research directions through which DRL algorithms' *adaptability* and *generalizability* can be enhanced and potentially make them applicable to a broad range of real-world problems.

## 1 Introduction

Humans are inherently capable of adapting quickly to changes in their environments and generalizing their past knowledge to tackle future *unseen* situations. Similarly, recent research in Artificial Intelligence (AI) is progressing at an impressive speed to create intelligent agents that can achieve human-level intelligence [Haenlein and Kaplan, 2019]. AI algorithms aim to possess an ability to *adapt* to unseen situations and *generalize* well across various domains. In recent years, Deep Reinforcement Learning (DRL) is gaining traction in the AI community due to a plethora of emerging approaches being used for developing efficient policies to solve complex problems.

AI research works actively use the notions such as *Domain Adaptation*; *Domain Generalization*; *Task Adaptation*; and *Task Generalization* as their learning goals. The simulated or real-world experiences are utilized in such research to attain these goals. However, there is a rare attempt to provide a clearer distinction among them. In this paper, we attempt to formulate Adaptation and Generalization in the context of Task and Domain and present the recent developments of state-of-the-art AI algorithms aimed at attaining these goals through DRL-based approaches.

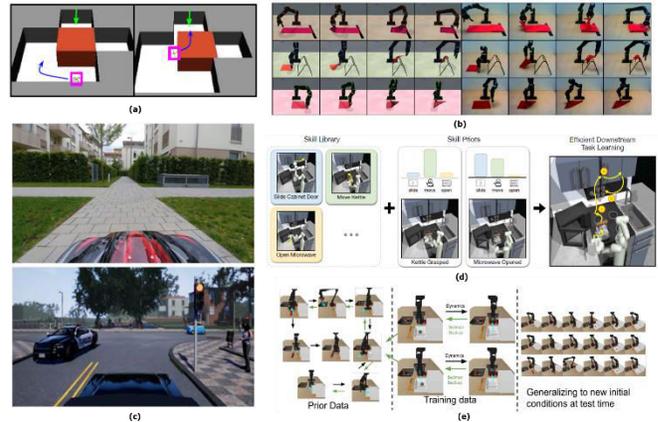

Figure 1: An illustration of complex problem settings with different learning goals: (a–c) Adaptation; (d, e) Generalization.

Fig. 1 shows some complex problems settings representing scenarios that are encountered in the real world. Fig.1 (a) demonstrates a navigation task in which the ant agent (magenta rectangle) has to reach the target destination (green arrow) given the path is obstructed by an object (orange block). The agent learns to push the block sideways and navigate to the target point using hierarchical reinforcement learning [Nachum et al., 2018]. Fig. 1(b) demonstrates a cloth-folding activity which at its base is a complex problem for a robot. [Matas et al., 2018] demonstrated a sim-to-real transfer mechanism to successfully perform this task by training in one domain (simulation) and applying it in another domain (real-world). Fig. 1(c) demonstrates that an autonomous vehicle is given an expert demonstration in a simulated environment which it learns to imitate in a real-world scenario [Codevilla et al., 2018]. Fig. 1(d) demonstrates an agent needs to learn a series of sub-tasks like moving kettle, opening the cabinet door, grasping kettle, and opening microwave to perform a given task of heating water. [Pertsch et al., 2020] accomplished to solve this task by the application of skill discovery in robotic manipulation. Fig. 1(e) demonstrates a complex set of scenarios where the robotic agent needs to perform actions like moving bottle, opening/closing drawer, grabbing spherical object without any well-defined instructions. [Singh et al., 2020] accomplished to generalize the agent to perform these tasks in unseen scenarios.

## 2 Preliminaries

**Definition 1** (Domain and Task). *A joint distribution $D_{XY}$ is called a domain, where $X = \{x_1, x_2, ..., x_m\}$ represents the input space and $Y = \{y_1, y_2, ..., y_n\}$ represents the output space ($m = n$ or $m \neq n$). $S \sim D_{XY}$ is the source domain, and $T \sim D_{XY}$ is the target domain. $S_X$ is defined as the marginal distribution on X in source domain. $T_X$ is defined as the marginal distribution on X in target domain. A learning function can be defined as $f: X \to Y$ whereas, the associated loss function can be defined as $l: Y \times Y \to [0, \infty)$. A set of tasks $t = \{t_1, t_2, ...\}$, such that $t_i \in Y$ can be drawn from either same or similar $D_{XY}$.*

**Definition 2** (Domain Adaptation (DA) and Task Adaptation (TA)). *DA can be defined as the prediction goal using optimized f when the conditions $S_X \neq T_X$ and $t^s = t^T$ hold true ($t^s$ represents task in source domain, and $t^T$ represents task in target domain). TA can be defined as the prediction goal using optimized f when the conditions $S_X = T_X$ and $t^s \neq t^T$ hold true.*

**Definition 3** (Domain Generalization (DG) and Task Generalization (TG)). *DG can be defined as the prediction goal using optimized f for unseen T. Here, unseen refers to T that is not seen during training time. TG can be defined as the prediction goal using optimized f for unseen $t^T$.*

### 2.1 Markov Decision Processes (MDPs)

RL settings are typically formulated using MDP having a tuple $(S, A, p_s, r)$ where S is a discrete or continuous state space; A is a discrete or continuous action space; $p_s$ is a transition function; and $r$ is a reward function. $p_s$ and $r$ are typically used to represent a model of the environment [Sutton et al., 1998].

### 2.2 Policy

A policy $\pi(a|s)$ is a mathematical function that defines the behavior of the agent by mapping a state $s_t \in S$ to an action $a_t \in A$. It can be either deterministic or stochastic but owing to the complex structure of the real-world yielding uncertainties in the outcomes promotes stochastic policies over deterministic ones while solving most of the real-world problems [Sutton et al., 1998].

## 3 DRL-based Approaches for Adaptation and Generalization

In this section, we present the various DRL-based approaches associated with their learning goals and describe them in detail in the following subsections. Fig. 2 shows various DRL-based approaches popularly used to attain adaptation and generalization in solving complex problems.

### 3.1 Hierarchical Reinforcement Learning (HRL)

Recently, various research works have applied HRL to solve the complex problems/tasks that constitute simpler related problems/tasks. [Vezhnevets et al., 2017] proposed a *transition policy gradient update* mechanism for a two-level hierarchical network inspired by [Dayan and Hinton, 1993]. It consists of a *manager* network that learns a latent state-space and sets goals, and a *worker* network that produces primitive actions based on the goals received from the manager. [Nachum et al., 2018] aimed at performing multi-level reasoning by proposing a method for learning a hierarchy of policies. It is divided as low-level policies - responsible for taking actions in the environment and high-level policies – responsible for planning the long-term decisions. It achieved higher sample efficiency by training the agent in an off-policy setting for complex tasks such as robotic locomotion and object manipulation in a simulated environment.

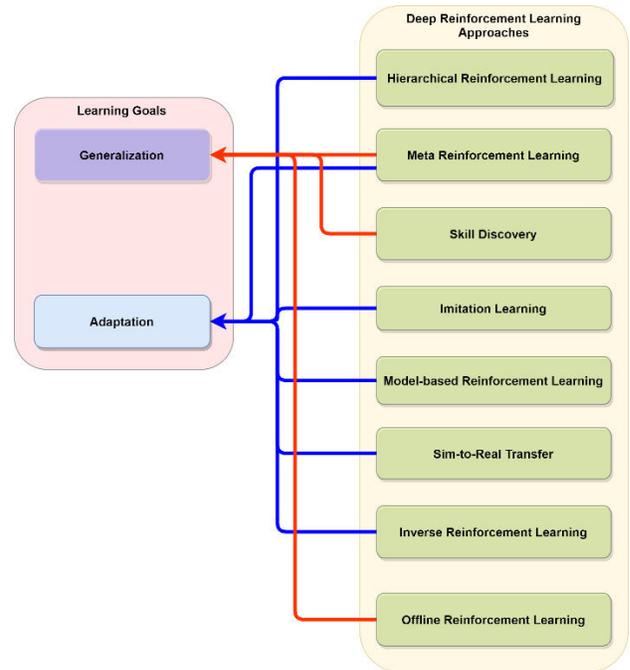

Figure 2: DRL-based approaches for adaptation and generalization.

It is critical to apply an appropriate objective for ensuring the usefulness of the constructed hierarchies. [Haarnoja et al., 2018] introduced a framework to automate this hierarchy construction process Also, they proposed an algorithm to train maximum entropy policies utilizing latent variables. [Nachum et al., 2019] attempted at solving the issue of choosing appropriate representations responsible for mapping the agent's states to the goals assigned by high-level policies. This was done by introducing a concept of sub-optimality to choose representation learning objective, which in turn can help find near-optimal hierarchical policies. [Hejna et al., 2020] proposed the usage of priors for capturing the learned knowledge by a low-level policy in a simpler environment that can be transferred to solve a similar prob-lem in a more complex environment. This low-level policy knowledge was transferred by leveraging an information-theoretic objective to perform effective high-level knowledge transfer. [Zhang et al., 2021a] tackled the problem of sparse reward environment by using a method that could grasp the abstractions in a self-supervised manner, consequently performing efficient intrinsic option discovery to obtain higher rewards for the task.

### 3.2 Meta Reinforcement Learning (Meta-RL)

Meta-RL is based on the notion of learning to learn [Thrun and Pratt, 1998]. Currently, it is widely used in the development of learning algorithms that are capable of generalization to unseen tasks/domains. [Li et al., 2018] made one of the first attempts to develop a model-agnostic algorithm as opposed to DG-specific models [Khosla et al., 2012; Ghifary et al., 2015], which was demonstrated on RL settings to perform improve DG ability. [Yoon et al., 2018] inspired by the application of Bayesian inference principle in RL [Houthooft et al., 2016], proposed the first Bayesian fast adaptation method to solve the overfitting problem prevailing in vanilla meta-learning algorithms [Finn et al., 2017] to adapt to unseen tasks by approximating the uncertainties in them using a meta-update loss calling it the Chaser loss. The primary motivation behind using Meta-learning methods is to achieve faster adaptation to unseen tasks without learning from scratch by utilizing past experience.

Correspondingly, meta-RL methods have started gaining popularity to achieve the same goal in an RL setting [Finn et al., 2017; Gupta et al., 2018a]. [Gupta et al., 2018b] improved the generalizing ability of these meta-RL algorithms by eliminating the bottleneck of task distribution specification. Although it assumes an unknown reward setting during meta-test time, it requires the environment dynamics to remain unchanged during both meta-train and meta-test times.

[Song et al., 2020] enabled the learning of adaptable policies for solving the task of robot locomotion in the real world. They introduced a Batch-Hill Climbing operator and integrated it with ES-MAML [Song et al., 2019] to achieve robustness to the noises in the real environment. [Najarro and Risi, 2020] proposed a widely different approach to perform meta-learning in RL environments by employing a Hebbian-inspired synaptic plasticity mechanism to help the neural networks in the learning algorithm adjust their weights for the tasks without using any reward function.

Update rule defined in the learning algorithm requires a re-iterative approach to tune in the parameters for each task. Meta-learning approach provides one way to learn the rules for learning. But the gradient-based update in standard meta-learning methods [Finn et al., 2017; Rusu et al., 2018] are computationally expensive because it requires evaluation of an update rule is preceded by running the meta-learning algorithm. Moreover, failing to hold the assumption of meta-learner's ability to learn after a given number of updates leads to degradation in generalization performance. [Flennerhag et al., 2021] proposed an algorithm to overcome two bottlenecks in the optimization of the meta-learner's objective: (i) meta-objective's constraint to the similar geometry as of the learner's; (ii) meta-objective's limited ability to generalize within the given number of steps and failing to incorporate the future dynamics. To eliminate (i), bootstrapping is employed in the algorithm to assimilate information about learning dynamics into the meta-objective; and to improve (ii), the meta-objective is defined in terms of minimizing distance to the bootstrapped target using KL-divergence $\mu^\pi(\tilde{x}, x) := KL(\pi_{\tilde{x}} \| \pi_x)$ [Yu et al., 2013].

### 3.3 Skill Discovery

A *skill* (or option) in the context of RL is defined as a latent-conditioned policy that can be trained to perform useful tasks in a sparse/unknown reward environment [Sutton et al., 1999]. [Achiam et al., 2018] proposed a variational inference-based option discovery method for training an agent to discover and learn skills through environment interaction without the need of maximizing the cumulative reward for a given task. [Eysenbach et al., 2018] proposed a reward-free skill learning method using an information-theoretic objective with a maximum entropy policy to keep the set of learned skills as diverse and discriminable as possible to maintain their usefulness. [Co-Reyes et al., 2018] introduced a novel hierarchical RL algorithm, which is capable of learning the skills in a continuous latent space (low-level). Also, they proposed a model that is capable of predicting the output of the learned skills. Such a capability can play a crucial role in solving more complex tasks at a higher level.

TG in an RL setting is a major challenge for achieving multi-task solving ability by an agent. [Petangoda et al., 2019] proposed a multi-task reinforcement learning method for learning disentangled skill embeddings [Higgins et al., 2017a] for dynamics and reward function. They also introduced two algorithms that are capable of learning multiple dynamics and multiple goal policies. Also, these algorithms are capable of performing new TG. [Sharma et al., 2020] proposed a low-level skill learning algorithm that uses an objective to embed the primitives (behaviors) into latent space. This low-level skill learning allows for better model-based planning for generalization over unseen tasks. [Campos et al., 2020] proposed an information-theoretic approach to perform skill (option) discovery. They also perform optimization of the information-theoretic objective used in previous approaches [Eysenbach et al., 2018; Sharma et al. 2020]. [Bagaria and Konidaris, 2020] proposed a deep skill chaining algorithm by unifying the neural networks with the skill chaining concept to scale up the ability of vanilla skill chaining [Konidaris and Barto, 2009] to high-dimensional tasks. [Pertsch et al., 2020] proposed an algorithm to learn a prior over skills to help the agent in exploring the skill space efficiently. Priors help in deciding which skill is more important to explore while performing a particular action. [Kim et al., 2021] have proposed an approach that uses a linearizer, to simplify the dynamics of an environment, doing so makes it easy to transition to the diverse states without external rewards; and an information-bottleneck option learning algorithm based on the information bottleneck framework [Tishby et al., 2000] to discover and learn skills.

### 3.4 Imitation Learning (IL)

IL approach allows an agent to learn by observing an expert demonstrating the required task [Hussein et al., 2017]. [Oh et al., 2018] proposed an off-policy actor-critic algorithm that focuses on learning to imitate the agent's past good experiences. It uses a replay buffer [Mnih et al., 2015] to store past experiences. The algorithm has been experimented with to perform well in several Atari games that require difficult exploration such as Montezuma's Revenge game as well as MuJoCo based control tasks by improving the performance of PPO [Schulman et al., 2017].

IL can be an effective way to solve the autonomous driving tasks by learning from the human driving demonstrations. [Codevilla et al., 2018] proposed a conditional imitation learning mechanism to overcome the myopic assumption that perception is sufficient for vehicle control [Bojarski et al., 2016]. Their mechanism trained the algorithm to incorporate the representation of the expert's intention along with the perception to perform efficient control in autonomous driving tasks.

The distributional shift has often been claimed to be a fundamental problem in IL [Ross and Bagnell, 2010]. BC approaches have succeeded in solving practical problems yielding good results [Bansal et al., 2019]. Distributional shift often gives rises to causal misidentification effect that leads to setting wrong correlation between the actions and their causes. [De Haan et al., 2019] proposed a method for learning the correct causal model to overcome the problem of causal misidentification. Their algorithm learns a mapping from causal graphs to the policies and then utilizes the knowledge of experts or real-world execution of the policies to select the correct policy for the target tasks. Imitation learning allows an agent to learn the preferences and goals of humans in a safe manner [Argall et al., 2009]. Providing high confidence bound on the agent's performance with the respect to the expert's demonstration is a relatively less-explored area. Much of the works in this area have used Bayesian Inverse Reinforcement Learning (Bayesian IRL) [Ramachandran and Amir, 2007] to measure both the reward uncertainty and the error over policy generalization [Brown et al., 2018]. However, (Bayesian IRL) typically suffers from the complex computation of MDPs leading to hindrance against the safety and efficiency of the model in unknown MDPs or high-dimensional problems. Providing preferences over goals is an intuitive action for humans [Akrour et al., 2011; Sadigh et al., 2017; Palan et al., 2019]. [Brown et al., 2020] proposed Bayesian Reward Extrapolation (Bayesian REX) method to show that preferences over demonstrations enable faster reward learning in high-dimensional problems, and also attains better efficiency in high-confidence performance bounds.

[Yu et al., 2020] proposed a Generative Intrinsic Reward driven Imitation Learning (GIRIL) algorithm that aims to attain better-than-expert performance from a one-life demonstration of the expert. This algorithm utilizes a VAE [Kingma and Welling, 2013] to generate diverse future states (decoder) and the corresponding action latent variables (encoder). This approach provides intrinsic rewards that in turn helps in performing better self-supervision exploration of the unseen environment. [Zhang et al., 2021b] considered an RL setting where the learned policies are suboptimal. It aims to solve the case where the expert demonstrations are insufficient in quality or quantity. They have proposed to use a confidence metric to measure the likelihood of the optimality of demonstrations. Their algorithm, Confidence-Aware Imitation Learning, learns both, a policy that can perform well and the confidence value for every state-action pair in the expert's demonstrations. It has outperformed other standard IL approaches for demonstrations with varying optimality.

### 3.5 Model-based Reinforcement Learning (Model-based RL)

Model-free RL has shown tremendous success for solving the tasks in a simulated environment [Schulman et al., 2015; Lillicrap et al., 2015; Silver et al., 2016]. However, model-free RL methods are generally considered highly sample-inefficient. It imposes a limitation on those methods in being applicable to real-world environments such as robotics. In contrast, Model-based RL is able to learn in a sample efficient manner by performing the policy optimization against the learned dynamics of the environment.

[Clavera et al., 2018] proposed a model-based meta-policy optimization (MB-MPO) algorithm to allow learning an ensemble of dynamics models and formulating the policy optimization as a meta-learning problem. MB-MPO exhibited low sample complexity on high-dimensional tasks showing its applicability to real-world robotics environments. [Nagabandi et al., 2019] proposed a model-based meta-reinforcement learning algorithm capable of performing online adaptation to dynamic environments. Model-based Meta-RL approach has shown evidence of improved sample efficiency than the model-free meta-RL methods [Duan et al., 2016; Finn et al., 2017]. The algorithm was experimented on in both simulated and real-world environments and outperformed the standard model-based meta-RL methods, model-free methods, and prior meta-RL methods.

[Kaiser et al., 2020] proposed a Simulated Policy Learning (SimPLE) algorithm to perform policy optimization within a learned model. It utilizes a convolutional encoder and decoder with video frames as input to develop a model utilizing the discrete latent variables. SimPLE was experimented on several Atari games and is shown to outperform the state-of-the-art Rainbow algorithm [Hessel et al., 2018].

Modern online RL methods have demonstrated great success in solving complex sequential decision-making problems [Schulman et al., 2015; Schulman et al., 2017; Hessel et al., 2018]. However, online RL algorithms typically require an iterative collection of experiences during training. This approach of a frequent collection of new data can be undesirable to several areas such as health [Murphy et al.,

2001], education [Mandel et al., 2014], robotics [Kalashnikov et al., 2018]; doing so can be prone to increase risk and cost of the system deployment. To handle this issue, [Matsushima et al., 2020] proposed a novel metric for evaluating the performance of RL algorithms called deployment efficiency, which records the number of increments in the data-collection policy during training. Their algorithm, Behavior-Regularized Model-Ensemble (BREMEN), aims to learn an ensemble of environment dynamics models and has been tested on high-dimensional continuous tasks exhibiting impressive deployment efficiency.

[Lee et al., 2020] proposed a context-aware dynamic model (CaDM) that can generalize across various environments composed of varying transition dynamics. CaDM utilizes an encoder to capture the contextual information that assists in adapting to the changing dynamics of the environment. The algorithm was experimented on several Open AI gym and MuJoCo environments and has been shown to outperform model-based meta-RL methods [Nagabandi et al., 2019].

### 3.6 Sim-to-Real Transfer

DA is a major challenge in RL [Bengio et al., 2013]. Input distributions in simulated RL environments differ from the real-world environment. However, the reward structure remains almost similar. Thus, it is desired to develop a technique that allows the agent to adapt to an unseen domain. [Higgins et al., 2017b] proposed an algorithm that can learn disentangled representation [Higgins et al., 2017a] of the environment's generative factors and utilizes this information to learn a robust source policy that can be transferred to the target domain i.e., real-world environment.

Much of the modern RL research involving robotics manipulation focuses on rigid object manipulations [Peters and Schaal, 2008; Gu et al., 2016]. However, deformable objects such as cloth folding [Miller et al., 2014] is an important day-to-day activity in human lives but is yet an underexplored area. One of the difficulties faced in this area is the large configuration spaces due to the large change in the configuration involved in manipulating the deformable objects. [Matas et al., 2018] presented an improvised version of DDPG [Lillicrap et al., 2015] to learn the policies for handling deformable objects such as towel folding and were able to transfer the policies to the real objects by employing domain randomization [Tobin et al., 2017]. Although the Sim-to-Real transfer of the learned policy is a convenient way to solve complex robotics tasks by training in the simulated domain and applying it in the real domain, complexities involved in the real domains such as contact dynamics, soft bodies, etc., hinders the seamless transfer of optimal policy. [Kaspar et al., 2020] proposed a novel method to overcome this limitation by constraining the joint and cartesian space of the simulated robot in the Operational Space Control framework (OSC) [Khatib, 1987] to learn a goal-conditioned policy. The proposed method was experimented on a Kuka LBR iiwa peg-in-the-hole task and was able to transfer the policy without any dynamics randomization [Peng et al., 2017].

Sim-to-real transfer of the policy sometimes results in the performance below expectation due to the unincorporated hidden information about the physical environment such as inaccurate dynamics. [Arndt et al., 2020] proposed a gradient-based meta-learning technique to adapt a broad spectrum of randomized dynamics in the simulated environment. The algorithm was demonstrated on a hockey-puck task by transferring the learned policy onto a real robotic arm which managed to hit the puck in a similar manner as in the simulated domain.

### 3.7 Inverse Reinforcement Learning (IRL)

IRL deals with understanding the objectives and rewards of an agent by observing its behavior. It infers a reward function from rollouts of expert policy and that leads to policy improvement and generalization [Piot et al., 2016]. Therefore, knowledge of the rewards is a primary goal of IRL, and apprenticeship is commonly used to acquire such policy learning from an expert. [Ramachandran and Amir, 2007] presented an algorithm for reward learning and apprenticeship learning. They have used the apprenticeship learning approach to utilize the prior knowledge and experiences from the actions of an expert to derive a probability distribution over the space of reward functions. To perform inference, they modified the Markov Chain Monte Carlo (MCMC) algorithm [Andrieu et al., 2003]. Their experimentations were very close to the true reward functions and showed better apprenticeship learning.

In complex tasks, appropriate knowledge of optimal reward function is the fundamental goal of an IRL approach. Generally, IRL approaches fail to adapt to the locally consistent constraints. Therefore, [Park et al., 2020] proposed a Constraint-based Bayesian Nonparametric Inverse Reinforcement Learning (CBN-IRL) for inferring task goals and constraints using Bayesian nonparametric IRL. It helps in recognizing the optimal reward function for complex manipulation tasks from a single demonstration. A task is divided into several subtasks such that a complex demonstration can be characterized by multiple subtasks with a corresponding set of local constraints. Doing so allows to model the local constraints and thereby resolve the complex demonstration in a simpler and single demonstration without losing generality.

### 3.8 Offline Reinforcement Learning (Offline-RL)

Offline-RL enables policy learning from a pre-collected dataset of trajectories/experiences for the tasks/domains where the agent is not allowed to interact with the environment. This approach aims to overcome the practical limitations of online RL such as dangerous or expensive interactions [Dulac-Arnold et al., 2019; Levine et al., 2020].

[Agarwal et al., 2020] demonstrated that recent off-policy algorithms [Dabney et al., 2018] can outperform the efficient policies [Mnih et al., 2015] trained on a DQN dataset. Also, they proposed a Q-learning algorithm that was able to achieve strong generalization in the offline setting and out-performed the algorithm proposed by [Dabney et al., 2018]. At its base, the algorithm ensured that optimal Bellman consistency was applied on several Q-value estimates obtained through random combinations. Poor off-policy evaluation causes inaccurate Q value estimation. Such iterative algorithms have two inherent weaknesses: Distribution shift and Iterative error. Distribution shift causes evaluation error, and it propagates to the evaluation step of current policy. Iterative error exploitation is caused by the error between Q estimate's errors, and it tends to overestimate at each step. Therefore, reusing the data causes amplification of such an error. [Brandfonbrener et al., 2021] proposed a one-step algorithm called offline approximate modified policy iteration (OAMPI) and demonstrated that it can outperform complex iterative algorithms on a variety of Offline-RL problems and at the very least set a strong baseline.

## 4 Discussions

Distributional shifts in the data between the training and testing domain is one of the primary causes behind the shortcomings in DRL research. It is found that most of the DRL-based approaches are still focused on achieving TA/DA. Therefore, it is necessary to further develop such approaches to achieve generalization over unseen tasks/domains.

Large and diverse datasets have allowed the application of Offline-RL algorithms to achieve generalization. However, it faces difficulty in scaling beyond laboratory settings because of the longer training time requirements and is mostly applied to problems in simulation or indoor robotics domain. Although there have been efforts for RL datasets development [Mo et al., 2018; Fu et al., 2020 Levine et al., 2020]. Still, there is a need to curate more advanced datasets that consists of real-world experiences, and also incorporates contextual information about the environment and other agents in it. Doing so will allow the researchers to develop DRL algorithms that can learn from the latent intricacies about the environment.

On the other hand, Meta-RL algorithms require lesser training time to achieve faster adaptability/generalizability. However, its current applications are limited to solving the tasks within indoor environments or controlled outdoor (closed-world) environments. The scaling of Meta-RL algorithms requires measuring their reliability before being deployed in an open-world environment. Similarly, skill discovery algorithms have also demonstrated the ability to perform generalization.

Higher levels of autonomy in Autonomous Vehicles (AVs) are typical examples of open-world environment. Despite great progresses made in DRL research, AVs are still an under-researched domain because current DRL approaches fail to incorporate the notion of safety for undesirable situations. Therefore, future research shall also focus on the development of safety-centric DRL algorithms.

## 5 Conclusion

DRL-based approaches are widely becoming popular with their algorithms for solving complex problems in both closed- and open-world environments. Adaptation and generalization are used as the learning goals to evaluate the algorithm's learnability. Adaptation refers to its ability to learn faster without re-training from scratch and generalization refers to its ability to extrapolate beyond the learned knowledge to tackle unseen environments. We found that DRL datasets play a crucial role in improving the generalizability of learning algorithms. However, state-of-the-art DRL dataset still lack in real-world based data, and this limits the adaptability and generalizability of the algorithms. Furthermore, there should be reliability measuring metrics incorporated in the algorithms to assess their safeness in real-world applications.

## Acknowledgments

This work was mainly supported by the Brain Pool Program through the National Research Foundation of Korea (NRF) funded by the Ministry of Science and ICT (NRF-2019H1D3A1A01071115). This work was partially funded by the Institute of Information & Communications Technology Planning & Evaluation (IITP) grant funded by the Korean government (MSIT) (No.2020-0-00056, to create AI systems that act appropriately and effectively in novel situations that occur in open worlds).

## References

[Achiam et al., 2018] Joshua Achiam, Harrison Edwards, Dario Amodei, and Pieter Abbeel. "Variational option discovery algorithms." *arXiv preprint arXiv:1807.10299*, 2018.

[Agarwal et al., 2020] Rishabh Agarwal, Dale Schuurmans, and Mohammad Norouzi. "An optimistic perspective on offline reinforcement learning." In *International Conference on Machine Learning*, pp. 104-114. PMLR, 2020.

[Akrour et al., 2011] Riad Akrour, Marc Schoenauer, and Michele Sebag. "Preference-based policy learning." In *Joint European Conference on Machine Learning and Knowledge Discovery in Databases*, pp. 12-27. Springer, Berlin, Heidelberg, 2011.

[Andrieu et al., 2010] Christophe Andrieu, Arnaud Doucet, and Roman Holenstein. "Particle markov chain monte carlo methods." *Journal of the Royal Statistical Society: Series B (Statistical Methodology)* 72(3): 269-342, 2010.

[Argall et al., 2009] Brenna D. Argall, Sonia Chernova, Manuela Veloso, and Brett Browning. "A survey of robot learning from demonstration." *Robotics and autonomous systems* 57(5): 469-483, 2009.

[Arndt et al., 2020] Karol Arndt, Murtaza Hazara, Ali Ghadirzadeh, and Ville Kyrki. "Meta reinforcement learning for sim-to-real domain adaptation." In *ICRA*, 2020.

[Bagaria et al., 2021] Akhil Bagaria, Jason K. Senthil, and George Konidaris. "Skill discovery for exploration and planning using deep skill graphs." In *ICML*, 2021.

[Bansal et al., 2018] Mayank Bansal, Alex Krizhevsky, and Abhijit Ogale. "Chauffeurnet: Learning to drive by imitating the best and synthesizing the worst". *arXiv preprint arXiv:1812. 03079*, 2018.

[Bengio et al., 2013] Yoshua Bengio, Aaron Courville, and Pascal Vincent. "Representation learning: A review and new perspectives." *IEEE TPAMI*, 35(8):1798-1828, 2013.

[Bojarski et al., 2016] Mariusz Bojarski, Davide Del Testa, Daniel Dworakowski, Bernhard Firner, Beat Flepp, Prasoon Goyal, Lawrence

D. Jackel, et al. "End to end learning for self-driving cars". *arXiv preprint arXiv:1604. 07316*, 2016.

[Brandfonbrener et al., 2021] David Brandfonbrener, William F. Whitney, Rajesh Ranganath, and Joan Bruna. "Offline rl without off-policy evaluation." In *NIPS,* 2021.

[Brown et al., 2020] Daniel Brown, Russell Coleman, Ravi Srinivasan, and Scott Niekum. "Safe imitation learning via fast bayesian reward inference from preferences." In *ICML*, pages 1165-1177, 2020.

[Campos et al., 2020] Campos, Víctor, Alexander Trott, Caiming Xiong, Richard Socher, Xavier Giró-i-Nieto, and Jordi Torres. "Explore, discover and learn: Unsupervised discovery of state-covering skills." In *ICML*, pages 1317-1327, 2020.

[Clavera et al., 2018] Ignasi Clavera, Jonas Rothfuss, John Schulman, Yasuhiro Fujita, Tamim Asfour, and Pieter Abbeel. "Model-based reinforcement learning via meta-policy optimization." In *Conference on Robot Learning*, pages 617-629. PMLR, 2018.

[Co-Reyes et al., 2018] John Co-Reyes, YuXuan Liu, Abhishek Gupta, Benjamin Eysenbach, Pieter Abbeel, and Sergey Levine. "Self-consistent trajectory autoencoder: Hierarchical reinforcement learning with trajectory embeddings." In *ICML*, pages. 1009-1018. PMLR, 2018.

[Codevilla et al., 2018] Felipe Codevilla, Matthias Müller, Antonio López, Vladlen Koltun, and Alexey Dosovitskiy. "End-to-end driving via conditional imitation learning." In *ICRA*, pages 4693-4700, 2018.

[Dabney et al., 2018] Will Dabney, Mark Rowland, Marc Bellemare, and Rémi Munos. "Distributional reinforcement learning with quantile regression." In *AAAI*, 2018.

[Dayan et al., 1992] Peter Dayan, and Geoffrey E. Hinton. "Feudal reinforcement learning." In *NIPS*, 1992.

[De Haan et al., 2019] Pim De Haan, Dinesh Jayaraman, and Sergey Levine. "Causal confusion in imitation learning." In *NIPS*, 2019.

[Deng et al., 2009] Jia Deng, Wei Dong, Richard Socher, Li-Jia Li, Kai Li, and Li Fei-Fei. "Imagenet: A large-scale hierarchical image database." In *CVPR*, pages 248-255, 2009.

[Duan et al., 2016] Yan Duan, John Schulman, Xi Chen, Peter L. Bartlett, Ilya Sutskever, and Pieter Abbeel. "Rl $^ 2$: Fast reinforcement learning via slow reinforcement learning." *arXiv preprint arXiv:1611.02779*, 2016.

[Dulac-Arnold et al., 2019] Gabriel Dulac-Arnold, Daniel Mankowitz, and Todd Hester. "Challenges of real-world reinforcement learning." *arXiv preprint arXiv:1904.12901*, 2019.

[Eysenbach et al., 2018] Benjamin Eysenbach, Abhishek Gupta, Julian Ibarz, and Sergey Levine. "Diversity is all you need: Learning skills without a reward function." *arXiv preprint arXiv:1802.06070*, 2018.

[Finn et al., 2017] Chelsea Finn, Pieter Abbeel, and Sergey Levine. "Model-agnostic meta-learning for fast adaptation of deep networks." In *ICML*, pages 1126-1135, 2017.

[Flennerhag et al., 2021] Sebastian Flennerhag, Yannick Schroecker, Tom Zahavy, Hado van Hasselt, David Silver, and Satinder Singh. "Bootstrapped meta-learning." *arXiv preprint arXiv:2109.04504*, 2021.

[Fu et al., 2020] Justin Fu, Aviral Kumar, Ofir Nachum, George Tucker, and Sergey Levine. "D4rl: Datasets for deep data-driven reinforcement learning." *arXiv preprint arXiv:2004.07219*, 2020.

[Ghifary et al., 2015] Muhammad Ghifary, W. Bastiaan Kleijn, Mengjie Zhang, and David Balduzzi. "Domain generalization for object recognition with multi-task autoencoders." In *ICCV*, pages 2551-2559, 2015.

[Gu et al., 2016] Shixiang Gu, Timothy Lillicrap, Ilya Sutskever, and Sergey Levine. "Continuous deep q-learning with model-based acceleration." In *ICML*, pages 2829-2838, 2016.

[Gupta et al., 2018a] Abhishek Gupta, Russell Mendonca, YuXuan Liu, Pieter Abbeel, and Sergey Levine. "Meta-reinforcement learning of structured exploration strategies." In *NIPS,* 2018.

[Gupta et al., 2018b] Abhishek Gupta, Benjamin Eysenbach, Chelsea Finn, and Sergey Levine. "Unsupervised meta-learning for reinforcement learning." *arXiv preprint arXiv:1806.04640*, 2018.

[Haarnoja et al., 2018] Tuomas Haarnoja, Kristian Hartikainen, Pieter Abbeel, and Sergey Levine. "Latent space policies for hierarchical reinforcement learning." In *ICML,* pages 1851-1860, 2018.

[Haenlein and Kaplan, 2019] Michael Haenlein and Andreas Kaplan. A brief history of artificial intelligence: On the past, present, and future of artificial intelligence. *California management review* 61(4):5-14, 2019.

[Hejna et al., 2020] Donald Hejna, Lerrel Pinto, and Pieter Abbeel. "Hierarchically decoupled imitation for morphological transfer." In *ICML*, pages 4159-4171, 2020.

[Hessel et al., 2018] Matteo Hessel, Joseph Modayil, Hado Van Hasselt, Tom Schaul, Georg Ostrovski, Will Dabney, Dan Horgan, Bilal Piot, Mohammad Azar, and David Silver. "Rainbow: Combining improvements in deep reinforcement learning." In *AAAI*, 2018.

[Higgins et al., 2017a] Irina Higgins, Loic Matthey, Arka Pal, Christopher Burgess, Xavier Glorot, Matthew Botvinick, Shakir Mohamed, and Alexander Lerchner. "beta-vae: Learning basic visual concepts with a constrained variational framework." In *ICLR*, 2017.

[Higgins et al., 2017b] Irina Higgins, Arka Pal, Andrei Rusu, Loic Matthey, Christopher Burgess, Alexander Pritzel, Matthew Botvinick, Charles Blundell, and Alexander Lerchner. "Darla: Improving zero-shot transfer in reinforcement learning." In *ICML*, pages 1480-1490, 2017.

[Houthooft et al., 2016] Rein Houthooft, Xi Chen, Yan Duan, John Schulman, Filip De Turck, and Pieter Abbeel. "Vime: Variational information maximizing exploration." In *NIPS*, 2016.

[Hussein et al., 2017] Ahmed Hussein, Mohamed Medhat Gaber, Eyad Elyan, en Chrisina Jayne. "Imitation learning: A survey of learning methods". *ACM Computing Surveys (CSUR)* 50(2): 1–35, 2017.

[Kaiser et al., 2019] Lukasz Kaiser, Mohammad Babaeizadeh, Piotr Milos, Blazej Osinski, Roy H. Campbell, Konrad Czechowski, Dumitru Erhan et al. "Model-based reinforcement learning for atari." *arXiv preprint arXiv:1903.00374*, 2019.

[Kalashnikov et al., 2018] Dmitry Kalashnikov, Alex Irpan, Peter Pastor, Julian Ibarz, Alexander Herzog, Eric Jang, Deirdre Quillen et al. "Scalable deep reinforcement learning for vision-based robotic manipulation." In *Conference on Robot Learning*, pages 651-673, 2018.

[Kaspar et al., 2020] Manuel Kaspar, Juan D. Muñoz Osorio, and Jürgen Bock. "Sim2real transfer for reinforcement learning without dynamics randomization." In *2020 IEEE/RSJ International Conference on Intelligent Robots and Systems (IROS)*, pages 4383-4388, 2020.

[Khatib, 1987] Oussama Khatib. "A unified approach for motion and force control of robot manipulators: The operational space formulation." *IEEE Journal of Robotics and Automation* 3(1): 43-53, 1987.

[Khosla et al., 2012] Aditya Khosla, Tinghui Zhou, Tomasz Malisiewicz, Alexei A. Efros, and Antonio Torralba. "Undoing the damage of dataset bias." In *ECCV*, pages 158-171, 2012.

[Kim et al., 2021] Jaekyeom Kim, Seohong Park, and Gunhee Kim. "Unsupervised Skill Discovery with Bottleneck Option Learning." *arXiv preprint arXiv:2106.14305*, 2021.

[Kingma and Welling, 2013] Diederik P. Kingma and Max Welling. "Auto-encoding variational bayes." *arXiv preprint arXiv:1312.6114*, 2013.

[Lee et al., 2020] Kimin Lee, Younggyo Seo, Seunghyun Lee, Honglak Lee, and Jinwoo Shin. "Context-aware dynamics model for generalization in model-based reinforcement learning." In *ICML*, pages. 5757-5766, 2020.

[Levine et al., 2020] Sergey Levine, Aviral Kumar, George Tucker, and Justin Fu. "Offline reinforcement learning: Tutorial, review, and perspectives on open problems." *arXiv preprint arXiv:2005.01643*, 2020.

[Li et al., 2018] Da Li, Yongxin Yang, Yi-Zhe Song, and Timothy M. Hospedales. "Learning to generalize: Meta-learning for domain generalization." In *AAAI,* 2018.

[Lillicrap et al., 2015] Timothy P. Lillicrap, Jonathan J. Hunt, Alexander Pritzel, Nicolas Heess, Tom Erez, Yuval Tassa, David Silver, and Daan Wierstra. "Continuous control with deep reinforcement learning." *arXiv preprint arXiv:1509.02971*, 2015.

[Mandel et al., 2014] Travis, Mandel Yun-En Liu, Sergey Levine, Emma Brunskill, and Zoran Popovic. "Offline policy evaluation across representations with applications to educational games." In *AAMAS*, 2014.

[Matas et al., 2018] Jan Matas, Stephen James, and Andrew J. Davison. "Sim-to-real reinforcement learning for deformable object manipulation." In *Conference on Robot Learning*, pages 734-743, 2018.

[Matsushima et al., 2020] Tatsuya Matsushima, Hiroki Furuta, Yutaka Matsuo, Ofir Nachum, and Shixiang Gu. "Deployment-efficient rein-

forcement learning via model-based offline optimization." *arXiv preprint arXiv:2006.03647*, 2020.

[Mnih et al., 2015] Volodymyr Mnih, Koray Kavukcuoglu, David Silver, Andrei A. Rusu, Joel Veness, Marc G. Bellemare, Alex Graves et al. "Human-level control through deep reinforcement learning." *Nature* 518(7540): 529-533, 2015.

[Mo et al., 2018] Kaichun Mo, Haoxiang Li, Zhe Lin, and Joon-Young Lee. "The adobeindoornav dataset: Towards deep reinforcement learning based real-world indoor robot visual navigation." *arXiv preprint arXiv:1802.08824*, 2018.

[Murphy et al., 2007] Susan A. Murphy, Linda M. Collins, and A. John Rush. "Customizing treatment to the patient: Adaptive treatment strategies." *Drug and alcohol dependence*, 2007.

[Nachum et al., 2018] Ofir Nachum, Shixian Gu, Honglak Lee, and Sergey Levine. "Data-Efficient Hierarchical Reinforcement Learning" *Proceedings of the 32nd International Conference on Neural Information Processing Systems*, pages 3307-3317, 2018.

[Nachum et al., 2019] Ofir Nachum, Shixian Gu, Honglak Lee, and Sergey Levine. "Near-optimal representation learning for hierarchical reinforcement learning." In *ICLR* 2019.

[Nagabandi et al., 2018] Anusha Nagabandi, Chelsea Finn, and Sergey Levine. "Deep online learning via meta-learning: Continual adaptation for model-based RL." *arXiv preprint arXiv:1812.07671*, 2018.

[Najarro and Risi, 2020] Elias Najarro and Sebastian Risi. "Meta-learning through hebbian plasticity in random networks." In *NIPS*, 2020.

[Palan et al., 2019] Malayandi Palan, Nicholas C. Landolfi, Gleb Shevchuk, and Dorsa Sadigh. "Learning reward functions by integrating human demonstrations and preferences." *arXiv preprint arXiv:1906.08928*, 2019.

[Park et al., 2020] Daehyung Park, Michael Noseworthy, Rohan Paul, Subhro Roy, and Nicholas Roy. "Inferring task goals and constraints using bayesian nonparametric inverse reinforcement learning." In *Conference on robot learning*, pages 1005-1014, 2020.

[Peng et al., 2018] Xue Bin Peng, Marcin Andrychowicz, Wojciech Zaremba, and Pieter Abbeel. "Sim-to-real transfer of robotic control with dynamics randomization." In *ICRA*, pages 3803-3810, 2018.

[Pertsch et al., 2020] Karl Pertsch, Youngwoon Lee, and Joseph J. Lim. "Accelerating reinforcement learning with learned skill priors." *arXiv preprint arXiv:2010.11944*, 2020.

[Petangoda et al., 2019] Janith C. Petangoda, Sergio Pascual-Diaz, Vincent Adam, Peter Vrancx, and Jordi Grau-Moya. "Disentangled skill embeddings for reinforcement learning." *arXiv preprint arXiv:1906.09223*, 2019.

[Peters and Schaal., 2008] Jan Peters and Stefan Schaal. "Reinforcement learning of motor skills with policy gradients." *Neural networks* 21(4): 682-697, 2008.

[Piot et al., 2016] Bilal Piot, Matthieu Geist, and Olivier Pietquin. "Bridging the gap between imitation learning and inverse reinforcement learning." *IEEE Transactions on Neural Networks and Learning Systems* 28(8): 1814-1826, 2016.

[Rakelly et al., 2019] Kate Rakelly, Aurick Zhou, Chelsea Finn, Sergey Levine, and Deirdre Quillen. "Efficient off-policy meta-reinforcement learning via probabilistic context variables." In *ICML*, pages 5331-5340, 2019.

[Ramachandran and Amir, 2007] Deepak Ramachandran and Eyal Amir. "Bayesian Inverse Reinforcement Learning." In *IJCAI*, pages 2586-2591, 2007.

[Rusu et al., 2018] Andrei A. Rusu, Dushyant Rao, Jakub Sygnowski, Oriol Vinyals, Razvan Pascanu, Simon Osindero, and Raia Hadsell. "Meta-learning with latent embedding optimization." *arXiv preprint arXiv:1807.05960*, 2018.

[Sadigh et al., 2017] Dorsa Sadigh, Anca D. Dragan, Shankar Sastry, and Sanjit A. Seshia. "Active preference-based learning of reward functions" In *Robotics: Science and Systems*, 2017.

[Schulman et al., 2015] John Schulman, Sergey Levine, Pieter Abbeel, Michael Jordan, and Philipp Moritz. "Trust region policy optimization." In *ICML*, pages 1889-1897, 2015.

[Schulman et al., 2017] John Schulman, Filip Wolski, Prafulla Dhariwal, Alec Radford, and Oleg Klimov. "Proximal policy optimization algorithms." *arXiv preprint arXiv:1707.06347*, 2017.

[Sharma et al., 2019] Archit Sharma, Shixiang Gu, Sergey Levine, Vikash Kumar, and Karol Hausman. "Dynamics-aware unsupervised discovery of skills." *arXiv preprint arXiv:1907.01657*, 2019.

[Silver et al., 2016] David Silver, Aja Huang, Chris J. Maddison, Arthur Guez, Laurent Sifre, George Van Den Driessche, Julian Schrittwieser et al. "Mastering the game of Go with deep neural networks and tree search." *Nature* 529(7587): 484-489, 2016.

[Singh et al., 2020] Avi Singh, Albert Yu, Jonathan Yang, Jesse Zhang, Aviral Kumar, and Sergey Levine. "Cog: Connecting new skills to past experience with offline reinforcement learning." *arXiv preprint arXiv:2010.14500*, 2020.

[Song et al., 2019] Xingyou Song, Wenbo Gao, Yuxiang Yang, Krzysztof Choromanski, Aldo Pacchiano, and Yunhao Tang. "Es-maml: Simple hessian-free meta learning." *arXiv preprint arXiv:1910.01215*, 2019.

[Song et al., 2020] Xingyou Song, Yuxiang Yang, Krzysztof Choromanski, Ken Caluwaerts, Wenbo Gao, Chelsea Finn, and Jie Tan. "Rapidly adaptable legged robots via evolutionary meta-learning." In *2020 IEEE/RSJ International Conference on Intelligent Robots and Systems (IROS)*, pages 3769-3776, 2020.

[Sutton and Barto, 1998] Richard S. Sutton, and Andrew G. Barto. "Introduction to reinforcement learning." MIT Press, 1998.

[Sutton et al., 1999] Richard S. Sutton, Doina Precup, and Satinder Singh. "Between MDPs and semi-MDPs: A framework for temporal abstraction in reinforcement learning." *Artificial intelligence* 112(1-2): 181-211, 1999.

[Thrun and Pratt, 1998] Sebastian Thrun, and Lorien Pratt. "Learning to learn: Introduction and overview." In *Learning to learn*, pages 3-17. Springer, Boston, MA, 1998.

[Tishby et al., 2000] Naftali Tishby, Fernando C. Pereira, and William Bialek. "The information bottleneck method." *arXiv preprint physics/0004057*, 2000.

[Tobin et al., 2017] Josh Tobin, Rachel Fong, Alex Ray, Jonas Schneider, Wojciech Zaremba, and Pieter Abbeel. "Domain randomization for transferring deep neural networks from simulation to the real world." In *2017 IEEE/RSJ international conference on intelligent robots and systems (IROS)*, pages 23-30, 2017.

[Vezhnevets et al., 2017] Alexander Sasha Vezhnevets, Simon Osindero, Tom Schaul, Nicolas Heess, Max Jaderberg, David Silver, and Koray Kavukcuoglu. "Feudal networks for hierarchical reinforcement learning." In *ICML*, pages 3540-3549, 2017.

[Yoon et al., 2018] Jaesik Yoon, Taesup Kim, Ousmane Dia, Sungwoong Kim, Yoshua Bengio, and Sungjin Ahn. "Bayesian model-agnostic meta-learning." In *NIPS*, 2018.

[Yu et al., 2013] Dong Yu, Kaisheng Yao, Hang Su, Gang Li, and Frank Seide. "KL-divergence regularized deep neural network adaptation for improved large vocabulary speech recognition." In *2013 IEEE International Conference on Acoustics, Speech and Signal Processing*, pages 7893-7897, 2013.

[Yu et al., 2020] Xingrui Yu, Yueming Lyu, and Ivor Tsang. "Intrinsic reward driven imitation learning via generative model." In *ICML*, pages 10925-10935, 2020.

[Zhang et al., 2021a] Zhang, Jesse, Haonan Yu, and Wei Xu. "Hierarchical reinforcement learning by discovering intrinsic options." *arXiv preprint arXiv:2101.06521*, 2021.

[Zhang et al., 2021b] Zhang, Songyuan, Zhangjie Cao, Dorsa Sadigh, and Yanan Sui. "Confidence-Aware Imitation Learning from Demonstrations with Varying Optimality." In *NIPS*, 2021.